\documentclass{article}
\usepackage{ijcai20}

\usepackage{times}
\usepackage{soul}
\usepackage{url}
\usepackage[hidelinks]{hyperref}
\usepackage[utf8]{inputenc}
\usepackage[small]{caption}
\usepackage{graphicx}
\usepackage{amsmath}
\usepackage{amsthm}
\usepackage{booktabs}
\usepackage{algorithm}
\usepackage{algorithmic}
\usepackage{subfigure}
\title{Temporal Pulses Driven Spiking Neural Network \\ for Fast Object Recognition in Autonomous Driving}

\author{
Wei Wang$^1$
\and
Shibo Zhou$^2$\and
Jingxi Li$^2$\and
Xiaohua Li$^2$\and
Junsong Yuan$^1$\And
Zhanpeng Jin$^1$
\affiliations
$^1$University at Buffalo, The State University of New York\\
$^2$Binghamton University, The State University of New York
\emails
wwang49@buffalo.edu,
szhou19@binghamton.edu
}

\begin{document}

\maketitle

\begin{abstract}
Accurate real-time object recognition from sensory data has long been a crucial and challenging task for autonomous driving. Even though deep neural networks (DNNs) have been successfully applied in this area, most existing methods still heavily rely on the pre-processing of the pulse signals derived from LiDAR sensors, and therefore introduce additional computational overhead and considerable latency. In this paper, we propose an approach to address the object recognition problem directly with raw temporal pulses utilizing the spiking neural network (SNN). Being evaluated on various datasets (including Sim LiDAR, KITTI and DVS-barrel) derived from LiDAR and dynamic vision sensor (DVS), our proposed method has shown comparable performance as the state-of-the-art methods, while achieving remarkable time efficiency. It highlights the SNN's great potentials in autonomous driving and related applications. To the best of our knowledge, this is the first attempt to use SNN to directly perform object recognition on raw temporal pulses.

% a self-created comprehensive temporal pulses dataset simulating LiDAR reflection in different road scenarios, DVS-barrel dataset. In addition, the proposed model was also evaluated on an object recognition dataset derived from dynamic vision sensor (DVS), which is an event-based camera showing great potential in autonomous driving. 
\end{abstract}

\section{Introduction}

%As is known to all, a difficult problem in machine vision is to judge the distance of object. 
One of the most challenging problems in autonomous driving is to detect objects at distance in 3D space with low latency. To provide accurate object localization in 3D space, it is necessary to provide the depth cue. 
%Traditional methods for 2D image processing usually perform poorly on this task as a single camera could not obtain accurate depth information \cite{lu2006single}. 
Generating depth map based on multi-camera can improve the depth accuracy, however, its high computing complexity makes it difficult to meet the real-time requirement \cite{park1998acquisition}. 
%Another awkward problem is that the optics camera is subject to great lighting conditions and the object’s accuracy is uncertain. 
%Another awkward problem for optics camera is its dependency on lighting conditions, which results in uncertain and unstable performance in extreme weather conditions.
Although radio detection and ranging (Radar) is immune to lighting variations, its short-wavelength properties neither allow the detection of small objects nor provide users precise object images. 
%LiDAR with short-wavelength and full-angle field characteristics can solve the above problems to a large extent. By means of the nature of LiDAR, LiDAR are becoming a reliable solution for high speed and precision object detection.
By using laser signals with relatively shorter wavelength than radio waves, light detection and ranging (LiDAR) system provides better range, higher spatial resolution and a larger field of view than Radar to help detect obstacles on the curves \cite{weitkamp2005lidar}. 
Compared with other sensors, LiDAR has significant advantages in detecting and recognizing objects at a long distance in a wide view, so that autonomous vehicles at high speed can take evasive actions in time.

Although a significant amount of research has been done for object classification from 3D point clouds, it remains an open question on how to detect and recognize objects with raw LiDAR temporal pulses.
LiDAR uses active sensors which emit their own energy sources for illumination, and detects/measures the reflected energy from the objects. The received pulses are then processed to generate high quality digital 3D point clouds for later tasks. Different learning algorithms, such as traditional feature extraction methods \cite{behley2013laser,wang2015voting,tatoglu2012point} and novel neural network methods \cite{li20173d,chen2017multi,oh2017object,kim2016robust,asvadideep} have been proposed for object detection and recognition based on point clouds data. Even though great success has been achieved, all these methods still rely on frame-based inputs and suffer from significant time delay and high computational overhead. 

In this work, we aimed at object recognition, a foundation task for many computer vision applications. Particularly in autonomous driving scenarios, given the tight power constraints and very high real-time requirements, we sought to propose a new alternative approach utilizing spiking neural networks (SNNs), which is fundamentally different from the mainstream methods based on different variants of conventional neural networks. Due to their biologically plausible nature, SNNs have remarkable advantages over conventional, mathematically rigorous NNs in terms of the superior time- and energy-efficiency \cite{ponulak2011introduction}, which come from several aspects: 1) SNN is hardware-friendly. Instead of relying on computation-intensive GPUs, SNNs allow efficient mappings to neuromorphic or specialized hardware (e.g., SpiNNaker, TrueNorth,), 
%with each logical neuron and synapse mapped to a hardware neuron and a hardware synapse respectively, 
breaking the computing bottlenecks of the von Neumann architecture and significantly reducing the energy consumption. 2) Information processing in SNNs is event-driven, meaning that the neurons are activated asynchronously and sparsely. According to the spin rate and data transmission rate of Velodyne LiDAR sensors, the collection of a data frame usually takes 50 - 200 ms, which is considerably long latency for autonomous driving tasks. Unlike conventional NNs, SNNs' asynchronous and sparse neuronal spiking behaviors will remove the constraints of frames and work from the arrival of the very first input, resulting in faster information propagation, smaller latency and lower power consumption. 3) Decisions can be derived without processing all the spikes, reducing the computational complexity significantly \cite{susi2018fns,rasshofer2005automotive}. 4) Our proposed SNN model has the ability to directly work with temporal pulses inputs, therefore multiple stages of the data digitization and pre-processing towards 3D point clouds are no longer needed. Hence, It is hypothesized that, leveraging the spike nature of data processing, SNN could process the LiDAR temporal pulses in a real-time manner.

Specifically, we have the following contributions:
\begin{itemize}
\item We applied SNN to directly process temporal pulse signals from LiDAR and proposed a novel SNN-based system utilizing adapted temporal coding for object recognition. The model was further extended to spiking convolution neuron network (SCNN) for different tasks.
%It includes the SNN, a directly implementation of temporal coding where information is encoded in spike times instead of spike rates. \hm{This sentence is meaningless}
% Based on the study existing SNN models and adapted an adequate model for this task. 
% The temporal pulse signals from LiDAR do not require any preprocessing and are directly used as the inputs of SNN.

\item We created a comprehensive temporal pulses dataset, ``Sim LiDAR'', which simulates LiDAR reflection of different road conditions and target objects in diverse noise environments. 
% Although some LiDAR datasets have already been available, they are preprocessed through DSP, such as the Udacity and KITTI datasets \cite{geiger2013vision}. 
Our dataset can meet the requirement of the proposed system and presents practical significance.

\item The performance of the proposed SNN-based object recognition system was evaluated with three different datasets. The experimental results showed that our system achieved remarkable accuracy along with superior time and computational efficiency, highlighting the potential of SNN in autonomous driving scenarios.

\end{itemize}

% The rest of the paper is organized as follows. Section \ref{sec:background} introduces the background. Section \ref{sec:snn_recog} describes the details of proposed SNN model with temporal coding. Experimental performance is evaluated and discussed in Section \ref{sec:evaluation}. We conclude the paper in Section \ref{sec:conclusion}.

\section{Background}
\label{sec:background}

% In our simulations, LiDAR system fires laser signals to the front of vehicle at certain frequencies.
% Raw temporal pulses are generated by LiDAR single photon detector array when it receives the reflected photons from the object. 
% In this way, depth information is recorded by the different delays of temporal pulses. 
% Spiking neural network circuit takes temporal pulses as inputs, and directly generate classification results. 

\subsection{Object Recognition Using 3D LiDAR}

\begin{figure}[tbp!]
\centering
\includegraphics[width=0.8\linewidth]{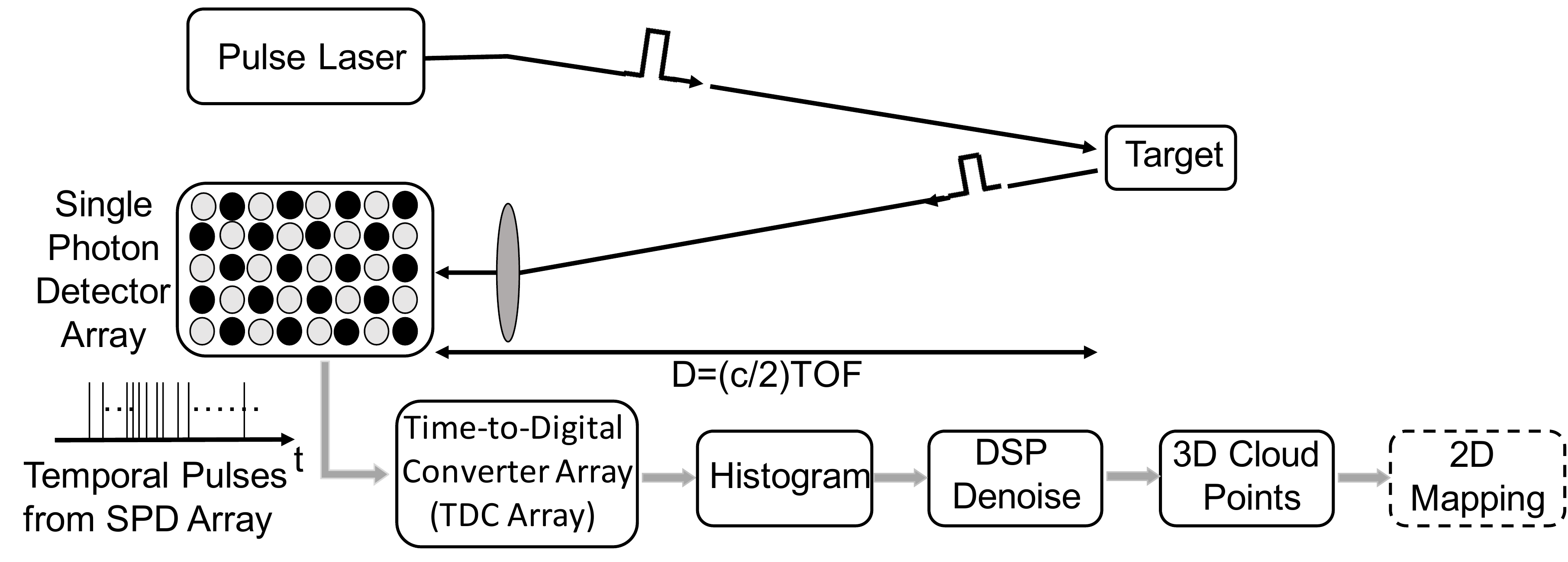}
\vspace{-10pt}
\caption{Generation of LiDAR point clouds}
\label{fig:lidar}
\vspace{-10pt}
\end{figure}

In recent years, LiDAR has been receiving increasing attention due to its high resolution and 3D monochromatic image on the object. 
%Firstly, a simple of overview of how the LiDAR works.
Generally, as illustrated in Figure \ref{fig:lidar}, the device emits laser pulses which move outwards in various directions until the signals reach an object, are reflected, and then return to the receiver. At the same time, an embedded processor saves each reflection points of a laser and generates 3D point clouds of the environment. Furthermore, the time interval between the moments when pulses leave the device and when they return to the LiDAR sensors are measured, which help determine the distance between a detected object and a LiDAR receiver. However, this whole process requires a series of transformations. When the signal returns to the receiver, the detector will generate a single photon detector array (SPDA) and record different temporal pulses information. Next, the time information corresponding to the temporal pulses needs to be converted into a digital form followed by histogram and DSP denoising. Finally, sometimes the 3D point clouds need to be further processed and transformed into 2D mapping. Based on the 3D point clouds data, the object detector can then try to differentiate different objects. In addition to 3D point clouds, the intensity of returning light, which is directly related to the reflectivity of the object, can also be used as a source for object detection. Nevertheless, the same transformation is required before the light intensity can be fed into the detection model. 
% After transformation, how to detect objects quickly and accurately, the post-processing system plays an important role in the detection of objects. 
A number of methods have been designed to perform accurate and efficient object detection. In general, these object detectors have followed two approaches: one is based on traditional methods such as hierarchical segmentation and sliding windows and the other combines NNs as the feature extractor or classifier. 
% Please refer to Section \ref{sec:related work} for more details.

% \begin{figure}[tbp!]
% \centering
% \includegraphics[scale=0.3]{fig/spiking_neuron.pdf}
% \caption{Spiking neuron model}
% \label{fig:spike_neuron}
% \end{figure}

\begin{figure}[tbp!]
\centering
\includegraphics[width=0.65\linewidth]{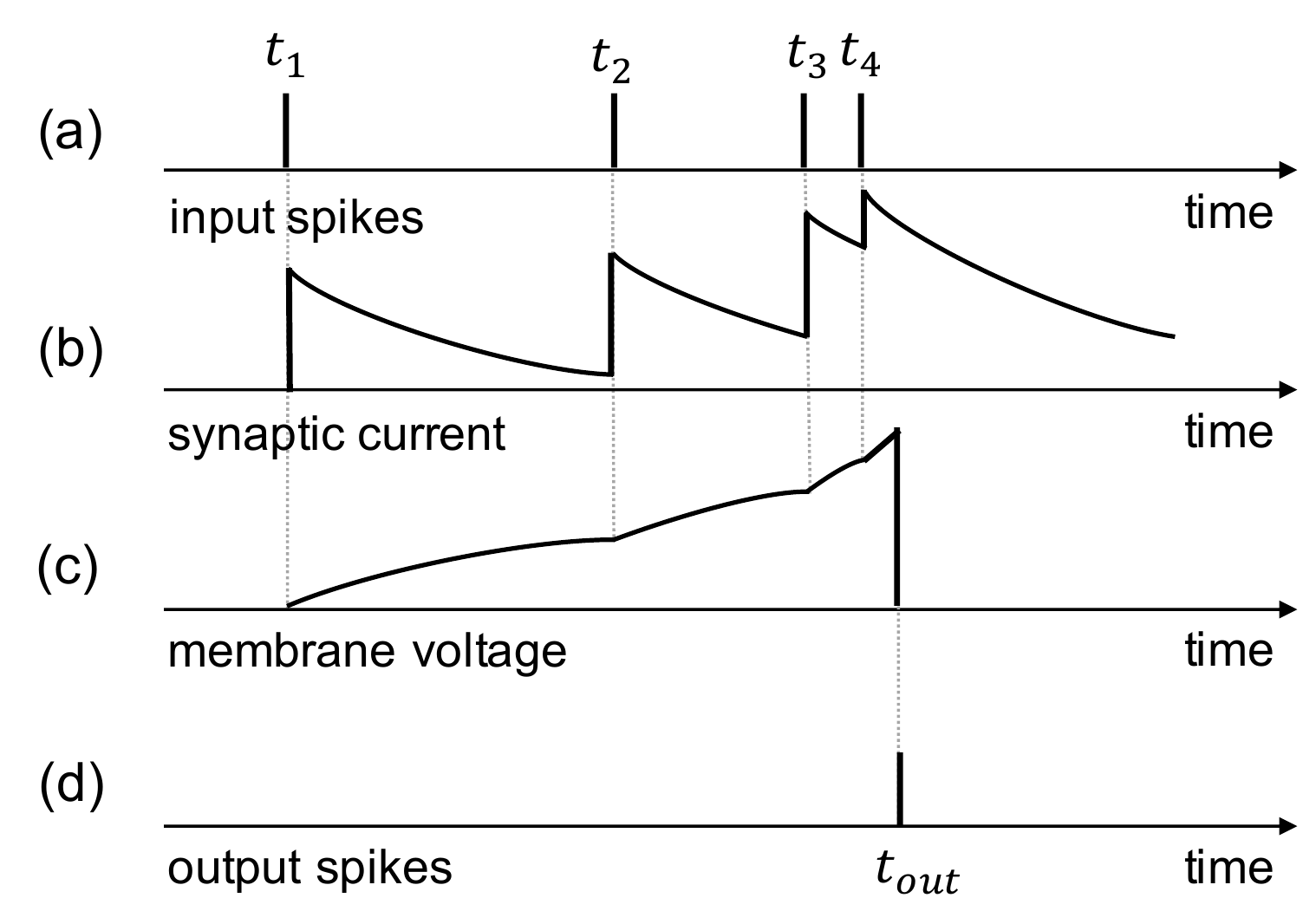}
\vspace{-10pt}
\caption{The working principle of the neuronal model. (a) Evoked by input spikes, (b) the synaptic current jumps and decays over time. (c) It then drives the membrane voltage potential of neuron cell to rise towards the firing threshold. (d) The neuron emits a spike whenever the threshold is crossed, in this case, at time $t_{out}$ after receiving 4 spikes with weights $\left \{ w_{1}, w_{2}, w_{3}, w_{4}\right \}$ at $\left \{ t_{1}, t_{2}, t_{3}, t_{4}\right \}$.} 
\label{fig:vmem}
\vspace{-10pt}
\end{figure}
\subsection{Spiking Neural Network Model}
\label{sec:snn_model}

%The artificial neural network (ANN) become more and more popular with the powerful computational ability in function estimation, complex pattern recognition, and classification. 
Spiking neural networks, which imitate biological neural networks by directly processing spike pulses information with biologically plausible neuronal models, are regarded as the third generation of artificial neural networks (ANNs) \cite{maass1997networks}. SNN's pulse processing mechanism enhances its capability to deal with spatio-temporal data, which constitutes the sensory data in real world, making it capable of processing temporal data without complex data pre-processing. 
%Given this unique capability, extensive efforts have been made to apply SNNs into different applications.
%Throughout their development, the ANN has evolved into the third generation: S (SNN) . 
% the pattern generator and controller for neuro-prosthetics systems \cite{ponulak2006resume}, adaptive robot path planning \cite{hwu2018adaptive}, obstacle recognition and avoidance by modeling and classifying spatio-temporal video data \cite{ge2017spiking}, 
%financial data prediction \cite{reid2014financial}, composer classification of a musical composition \cite{prasad2015composer}, 
% spatio- and spectro-temporal brain data mapping \cite{kasabov2014neucube}, etc. 
% At the same time, ROLLS microprocessor along with a Dynamic Vision Sensor \cite{qiao2015reconfigurable} and TrueNorth chip \cite{merolla2014million} has demonstrated superior performance in detection and classification. 
In SNNs, neurons communicate with spikes or action potentials through layers of network. When a neuron's membrane potential reaches its firing threshold, the neuron will emit a spike and transmit it to other connected neurons. 
Similar to conventional neural networks, SNN topology can be roughly classified into three categories: feedforward networks, recurrent networks and hybrid networks \cite{ponulak2011introduction}. 
In this work, we adopted the feedforward network as well as an variant with spiking convolution layers for tasks with different complexities. 
% Figure \ref{fig:snn} shows an example of fully-connected feedforward SNN, it includes three layers as input layer, hidden layer, and output layer. Number of hidden layers can be more than one for more powerful and complex neural networks. 
%Figure \ref{fig:spike_neuron} illustrates a spiking neuron model, which involves accumulation and thresholding operations. 
We also used the non-leaky integrate and fire (n-LIF) neuron with exponentially decaying synaptic current kernels \cite{mostafa2018supervised}.
The neuron's membrane dynamic is described by:
\begin{equation}
\centering
\frac{dV^{j}_{mem}(t)}{dt} = \sum\limits_{i}\omega _{ji}\sum\limits_{r}\kappa(t-t^{r}_{i})\vspace{-5pt}
\end{equation}
where $V_{mem}^{j}$ is the membrane potential of neuron $j$. $w_{ji}$ is the weight of the synaptic connection from neuron $i$ to neuron $j$ and $t_i^r$ is the time of the $r^{th}$ spike from neuron $i$. 
$\kappa$ is the synaptic current kernel function as given below:
\begin{equation}
\resizebox{.91\linewidth}{!}{$
\displaystyle
\centering
\kappa(x)=\Theta(x)\exp(-\frac{x}{\tau_{syn}}),\  \text{where} \ \Theta(x) = 
\begin{cases}
1 & \text{if $x\geq0$}\\
0 & \text{otherwise}
\end{cases}
$}
\end{equation}
where $\tau_{syn}$ is the only time constant. Once the neuron receives a spike, the synaptic current will jump instantaneously, then decays exponentially with time constant $\tau_{syn}$. Figure \ref{fig:vmem} shows how this SNN works. The spike is transformed by the synaptic current kernel and a neuron is only allowed to spike once unless the network is reset or a new input pattern is presented \cite{mostafa2018supervised}. 

\section{Object Recognition with SNN}
\label{sec:snn_recog}

\begin{figure}[tbp!]
\centering
\includegraphics[width=\linewidth]{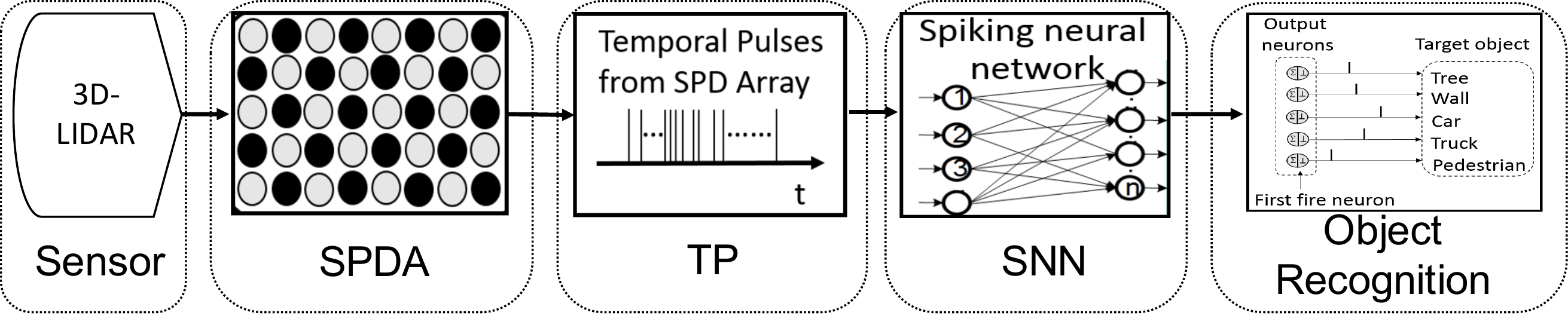}
\caption{Flow diagram of SNN-based object recognition}
\label{fig:pipeline}
\vspace{-10pt}
\end{figure}

Figure \ref{fig:pipeline} shows the workflow of our proposed SNN-based object recognition system for temporal pulses signals. Compared with the standard procedure in Figure \ref{fig:lidar}, as soon as one photon reflected from the target is received by the detector, SNN could directly process the raw pulse from SPDA, thus achieving lower latency. This pipeline generally consists of three parts: 1) the acquisition of temporal pulses, which contains object information, from SPDA; 2) the processing of temporal pulses signal using SNN with temporal coding; 3) object recognition and classification with the SNN model.

\subsection{SNN with Temporal Coding}

The time delay of temporal pulses from LiDAR's SPDA carries object information and the pulses from different objects exhibit diverse dynamics. To achieve satisfactory latency and accuracy, a proper coding mechanism of the pulse sequences plays a crucial role for computer simulation. 
% Therefore, how to implement simulation through coding plays a crucial role for this task. 

Usually, spike counts or rates within a time window are adopted for coding, but the spike counts are discrete values, which makes the training a great challenge. To avoid the problem, we chose the more precise, continuous spike times as the information-carrying quantities, in other words, the temporal information is used as a coding form of SNN. 

Existing temporal coding algorithms still face some limitations. The SpikeProp algorithm \cite{bohte2002error}, which describes the cost function in terms of the difference between the desired and actual spike times, is limited to learning a single spike. Supervised Hebbian learning \cite{legenstein2005can} and ReSuMe \cite{ponulak2010supervised}, are primarily suitable for the training of single-layer networks, but can hardly perform more complex computation. To address these issues, we adopted a simulation method for SNN with temporal coding that only relies on simple neural and synaptic dynamics instead of complex and discontinuous dynamics of spiking neurons \cite{mostafa2018supervised}. The original work only utilized a simple feedforward network with one hidden layer, and furthermore, all the inputs were binarized to provide more distinct temporal separation between spikes. This significantly limits the model's generalization capability for complex applications like autonomous driving. To this aim, we improved the temporal coding scheme to enable the processing of all continuous temporal values by normalization and re-scaling. Hence, we not only avoid the discreteness of spikes, but also extend it naturally to multi-layer networks with spiking convolution on complex tasks. As a result, our proposed SNN with temporal coding realized the direct processing of temporal pulses signal from LiDAR.

\subsection{SNN Processing Temporal Pulses Signal}

SNN with temporal coding can effectively perform the processing of temporal pulses signal in real time. The activation function derived from the non-leaky integrate and fire neurons expresses the relationship between the input spike times and the time of the first spike at output as below:
\begin{equation}
\centering
exp(t_{out}) = \frac{\substack{\sum_{i\in C}\omega_{i}exp(t_{i})}}{\substack{\sum_{i\in C}\omega_{i}}-threshold}
\end{equation}
where $t_{out}$ is the neuron's response time. $t_{i}$ is the firing time of $i$-th source neuron. $w_{i}$ is the weight corresponding to $i$-th source neuron. $C=\left \{ i: t_{i}\ <\ t_{out}\right \}$, $threshold$ is set to 1. 

% \begin{figure}[bp!]
% \vspace{-15pt}
% \centering
% \includegraphics[width=0.8\linewidth]{fig/snn.pdf}
% \vspace{-10pt}
% \caption{SNN network for object recognition}
% \label{fig:implementation}
% \vspace{-10pt}
% \end{figure}

\begin{figure}[tbp!]
\centering
\includegraphics[width=0.9\linewidth]{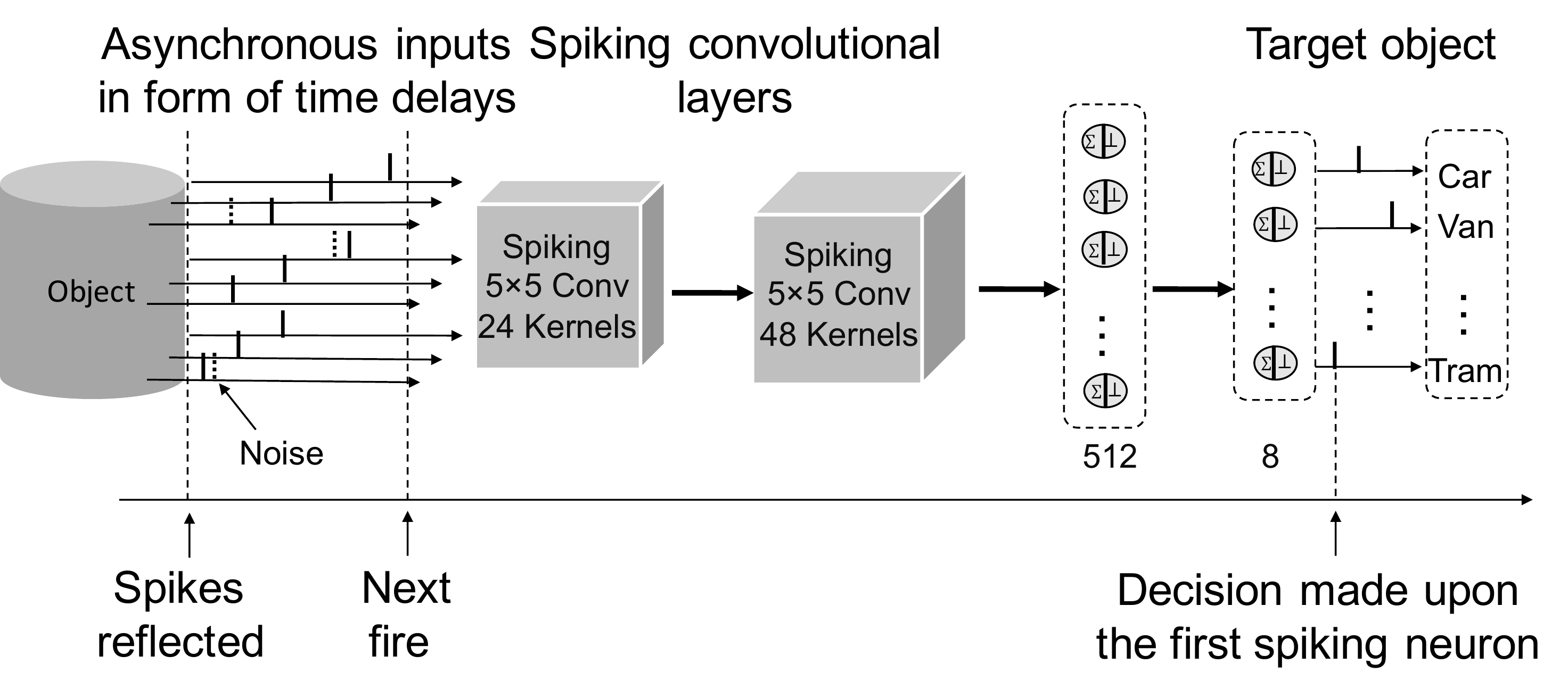}
\vspace{-10pt}
\caption{Spiking CNN network for object recognition}
\label{fig:scnn}
\vspace{-10pt}
\end{figure}

In order to handle tasks with different complexities, we implemented both feed forward SNN and deeper spiking convolution neural network (SCNN). The architecture of SCNN is shown in Figure \ref{fig:scnn}, which also demonstrates the mechanism of SNN processing temporal pulses. The original time is set as zero, after that, SNN will take in spikes sequentially according to their arriving time. Whenever a pulse comes in, the spiking neuron will keep accumulating the weighted value and comparing with the threshold until the accumulation of a set of spikes can fire the neuron. Once the neuron spikes, it would not process any further pulse unless being reset or presented with a new input pattern. That being said, the recognition result is made at the time of the first spike among output neurons. Therefore, not all spikes are required for the SNN to finish a pattern recognition, allowing the SNN to have faster responses. The standard back propagation technique can be used to train the weights. The spiking convolution layers work in a similar way as the traditional CNN but being equipped with spiking kernels to process the pulse signals.

%  Furthermore, We set the classification through the first fire neuron of output layer, which is beneficial to the acceleration of results. So, the trained SNN could process directly temporal pulses signal from LiDAR and implement object detection with high accuracy and real time.

\section{Experiments and Evaluations}
\label{sec:evaluation}

\subsection{Datasets}
To investigate the effectiveness of our proposed model in autonomous driving scenarios, we evaluated the model on three different datasets: 1) Sim LiDAR, a self-generated comprehensive temporal pulses dataset simulating LiDAR reflection in different road scenarios; 2) KITTI 3D object detection benchmark \cite{Geiger2012CVPR}, a real-world computer vision benchmark derived from autonomous driving platforms; 3) DVS\_barrel \cite{orchard2015hfirst}, an object recognition dataset derived from dynamic vision sensors (DVS).

\subsubsection{Sim LiDAR}

\begin{figure}[tbp!]
\centering
\includegraphics[width=0.9\linewidth]{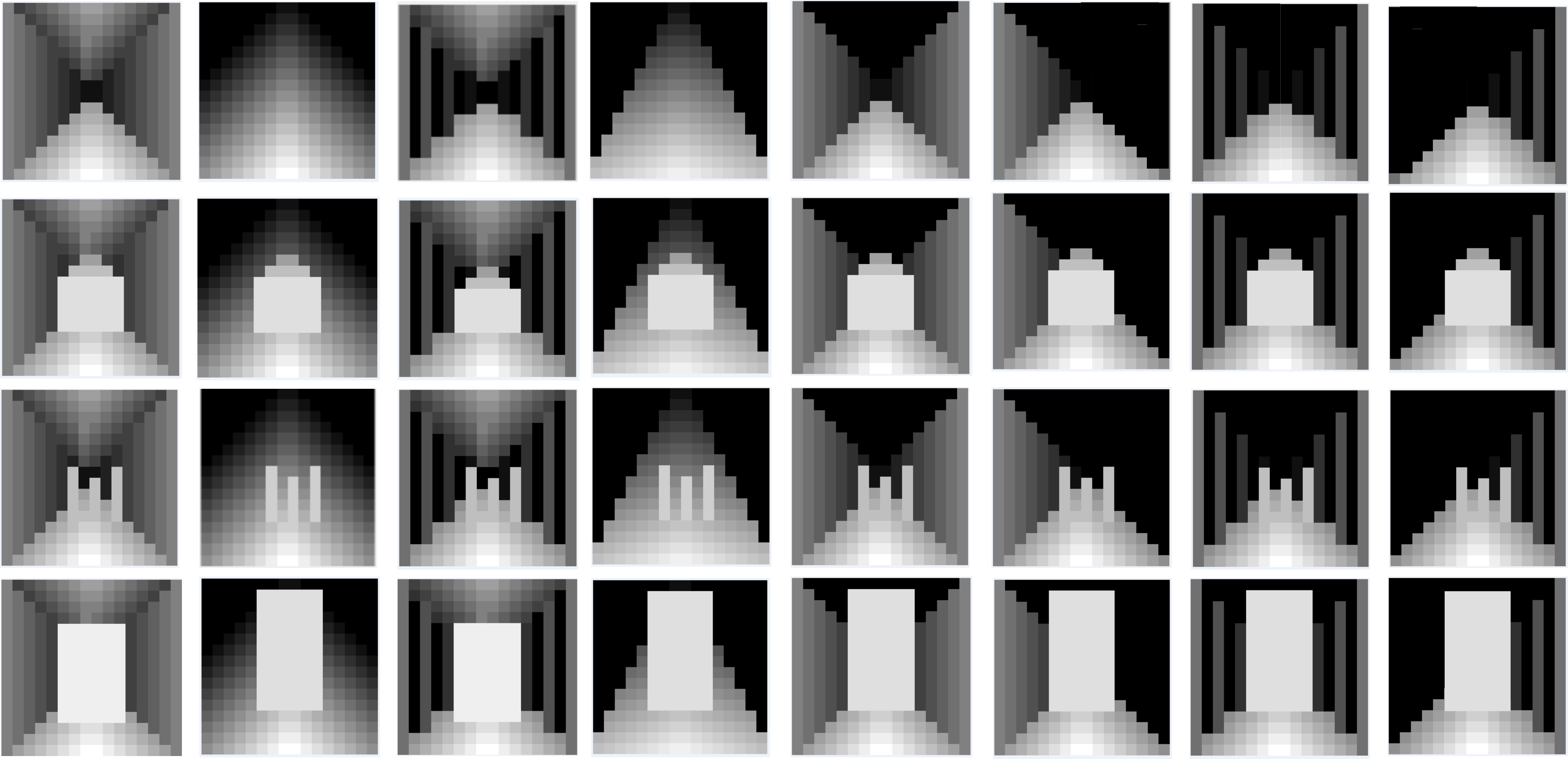}
\caption{Sim LiDAR covers different objects and road conditions}
\label{fig:sim_lidar}
\vspace{-10pt}
\end{figure}

\begin{figure}[tbp!]
    \centering
    \subfigure[$0-0.1$]{
        \includegraphics[scale=0.27]{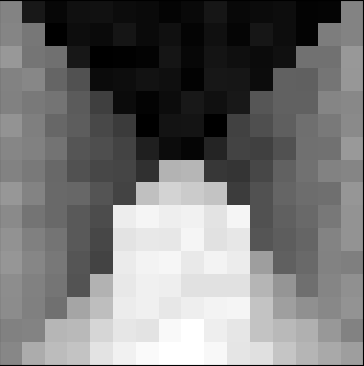}
        \label{fig:noise0.1}
    }\quad 
    \subfigure[$0-0.2$]{
        \includegraphics[scale=0.27]{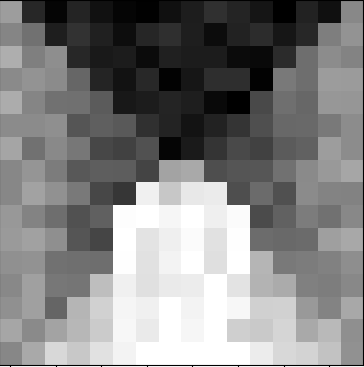}
        \label{fig:noise0.2}
    }\quad 
    \subfigure[$0-0.33$]{
        \includegraphics[scale=0.27]{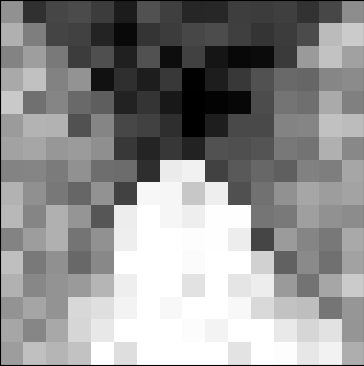}
        \label{fig:noise0.33}
    }\quad 
    \subfigure[$0-0.5$]{
        \includegraphics[scale=0.27]{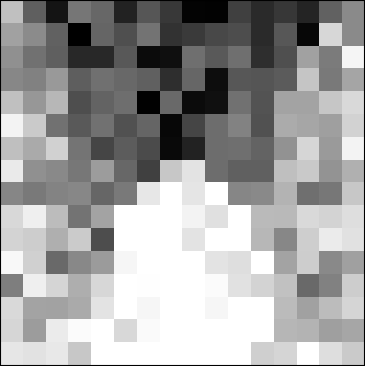}
        \label{fig:noise0.5}
    }
    \vspace{-10pt}
    \caption{A pattern with different ranges of noise}
    \label{fig:noise}
\vspace{-10pt}
\end{figure}

Our proposed system is expected to directly work with sensory temporal pulses data (e.g., LiDAR). However, instead of using raw data, most existing LiDAR datasets, such as Udacity and KITTI, are composed of either frame images or point clouds data, which is the elaboration product of LiDAR pulses after a series of processing. In view of this, we created a comprehensive temporal pulses database simulating LiDAR reflection in different road scenarios, named Sim LiDAR, to meet the experimental requirements and the actual situation. To generate the dataset, we used the Velodyne VLP-16 as a simulation LiDAR. The LiDAR emits some pulses at time $t$, the pulses will then be reflected by the obstacle and received by the sensor cell later with different time delays. So, all pulses own a time delay property, which is utilized to form the time delay array. It usually involves two cases: for the same object, due to the irregular surface, different parts would cause different time delays of the pulses; for different objects, because of the invariant vehicle speed of time, farther ones always result in larger delays, and vice versa. Based on the above rules, the source data is generated using MATLAB. The dataset is visualized in Figure \ref{fig:sim_lidar} based on LiDAR's 3D projection to demonstrate test scenarios, where LiDAR is located at the bottom center of the view and different gray scales represent different time delays. Each row of the data samples (sample size of 16$\times$16) corresponds to a different object (none, car, pedestrians and truck) in a variety of road conditions (from left to right: tunnel, open road, lower bridge, upper bridge, road with two side walls, road with one side wall, road with street lamps on both sides and with street lamps on one side), making for 32 classes. Augmentation can be further made for each class by shifting the values of time delay and adding noise randomly. Thus, the dynamic characteristics of different objects can be sufficiently simulated and a variety of data samples with diverse patterns were created. 
%The target objects, combined with different road conditions, form the combinations of 24 different situations. 
% Because the sample size is 16$\times$16, to avoid severe overlaps, it can not contain too many kinds of objects. So we only designed eight different combinations for each target object, making for the 24 situations that contains a target. 

%  And then if we add different noises to each pattern, it will produce scenes with different noise. Considering the particularity of the dataset, the data generated by this method has two advantages. First of all, each class has a variety of patterns. Next, different noise scenarios are generated. Therefore, the data-set satisfies the application requirements for testing the performance of our system.

\paragraph{Noise Injection on Sim LiDAR}

% For the Noise: First of all, the noise here is not a real noise, it simply refers to the object is detected without a label or suspended matter in the air.

To simulate the real-life sensor and ambient noise, a random 16$\times$16 noise matrix was generated with the uniform distribution and added to each sample. During this process, data samples with different noise levels can be obtained by limiting the amplitude and range of noise to be injected. Figure \ref{fig:noise} shows the impact of different noise levels on a particular pattern.
% (i.e., the upper bound of noise ranges from 0.1 to 0.5). 
% From Figure \ref{fig:noise0.1} the range of noise is from 0 to 0.1, and the noise is not a great influence on a pattern. 
% From left to right, the upper bound of noise ranges from 0.1 to 0.5. 
%As severe distortions appear in Figure \ref{fig:noise0.33} and \ref{fig:noise0.5}, 
It can be observed that, the higher noise level is applied, the more severe distortions of the images are resulted, and the less information can be retrieved from objects. Based on the proposed Sim LiDAR dataset with noise injection, we are able to evaluate the  robustness and resistance of our system to noise. The final dataset contains 3000 training samples and 600 testing samples covering all 32 categories.

\subsubsection{KITTI}

\begin{figure}[tbp!]
\centering
\includegraphics[width=0.9\linewidth]{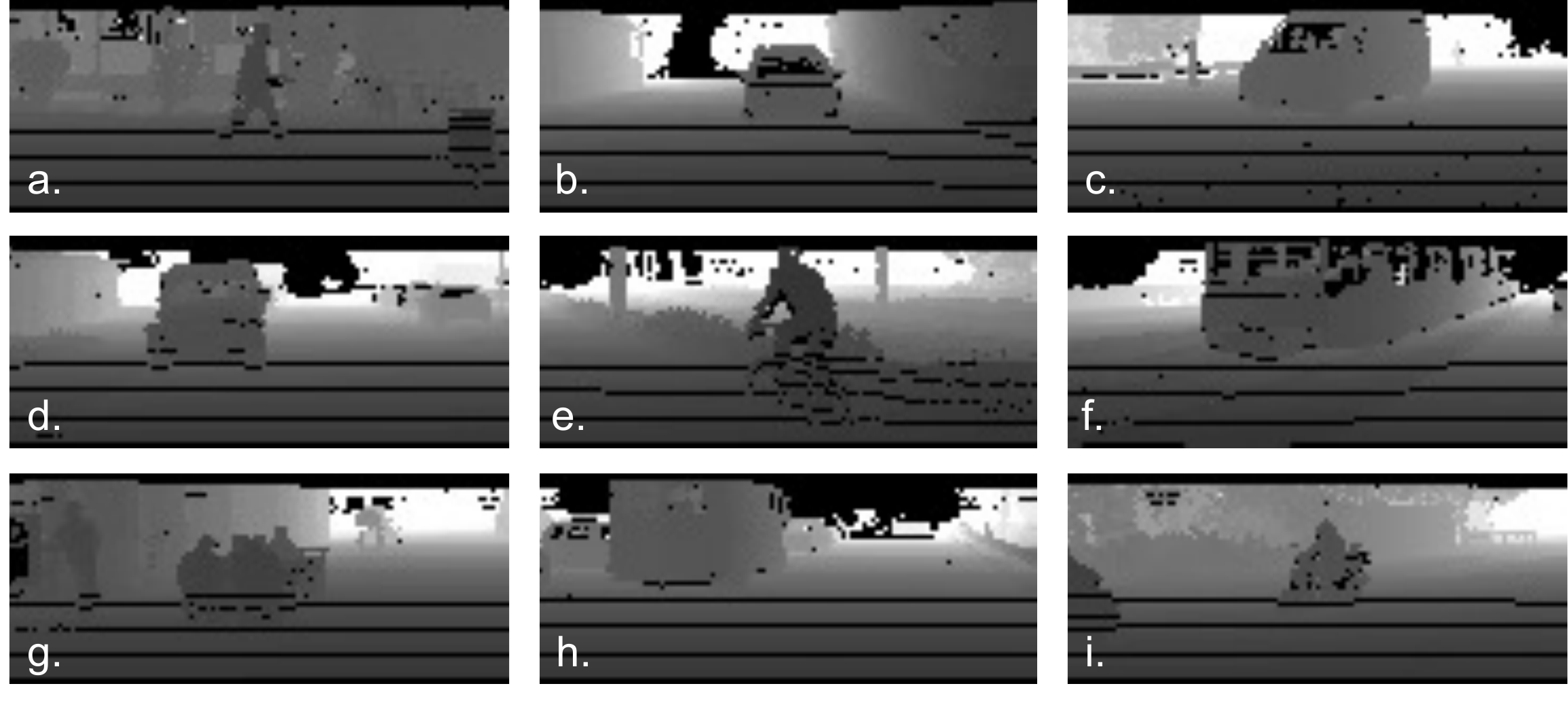}
\caption{Transformed KITTI dataset including 8 classes of objects: `Pedestrian' (a), `Car' (b), `Van' (c), `Truck' (d), `Cyclist' (e), `Tram' (f), `Person\_sitting' (g) and `Misc' (h, i).}
\label{fig:kitti}
\vspace{-10pt}
\end{figure}

The effectiveness of our proposed SNN-based system was also evaluated using the KITTI dataset \cite{Geiger2012CVPR}, which is a set of challenging real-world computer vision benchmarks captured based on the autonomous driving platform AnnieWAY. We utilized the KITTI 3D object detection benchmark, specifically the point clouds data collected by the Velodyne HDL-64E rotating 3D laser scanner, providing 7481 labeled samples. However, the provided point clouds data can not be directly used due to the following reasons: 1) point clouds data is derived after sophisticated processing, while the SNN model can directly handle the raw temporal pulses; 2) all label annotations (location, dimensions, observation angle, etc.) are provided in camera coordinates instead of the Velodyne coordinates. Therefore, the annotation data should be converted into the Velodyne coordinates and the point clouds data should be processed with temporal coding so as to reconstruct the original temporal pulses.

\begin{figure}[tbp!]
\centering
\begin{minipage}{.55\linewidth}
  \centering
  \includegraphics[width=\linewidth]{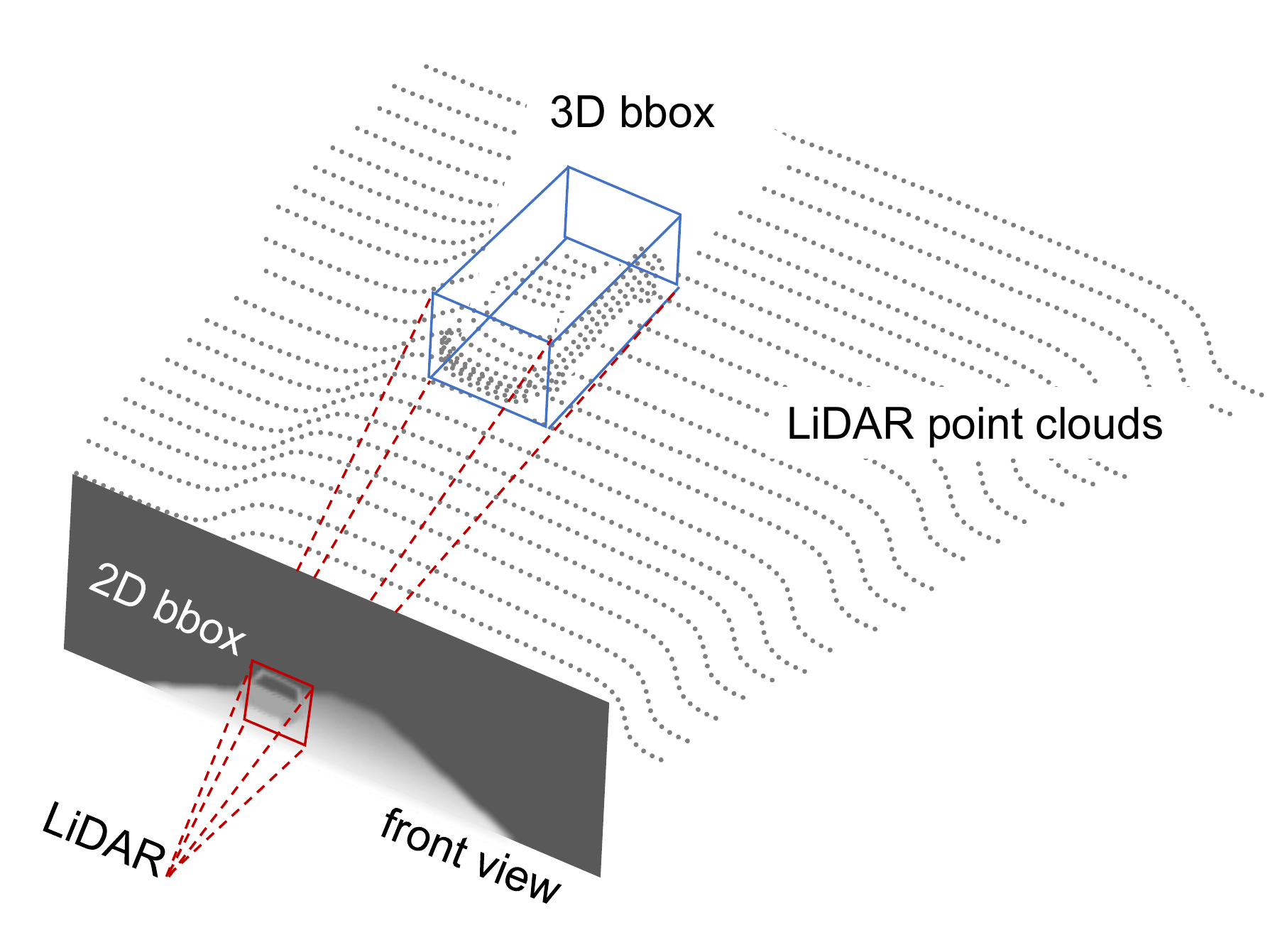} \vspace{-10pt}
  \captionof{figure}{KITTI transformation}
  \label{fig:kitti_process}
\end{minipage}%
\begin{minipage}{.45\linewidth}
  \centering
  \includegraphics[width=\linewidth]{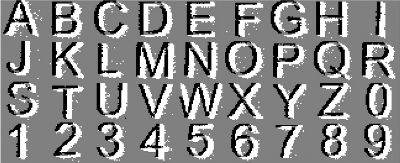}
  \vspace{-10pt}
  \captionof{figure}{DVS\_barrel dataset}
  \label{fig:dvs}
\end{minipage}
\vspace{-15pt}
\end{figure}

\paragraph{Transformation on KITTI}
% \begin{figure}[tbp!]
% \centering
% \includegraphics[width=0.6\linewidth]{fig/kitti_process.pdf}
% \caption{Transformation of KITTI Point Clouds}
% \label{fig:kitti_process}
% \vspace{-10pt}
% \end{figure}
The point clouds data provides the location ($x, y, z$ coordinate values in 3D space) and the corresponding reflectivity ($r$) of a series of points. However, to reconstruct the temporal pulses array, the point clouds need to be mapped into an expanded front view, whose size is determined by the resolution of the LiDAR sensor. The transformation of point clouds follows the equations below:
\begin{equation}
\centering
x_{front} = \left \lfloor arctan2(-y,x)\cdot \frac{180}{\pi\cdot Res_{h}} \right \rfloor
\end{equation}

\begin{equation}
\centering
y_{front} = \left \lfloor -arctan2(z,d)\cdot \frac{180}{\pi\cdot Res_{v}} \right \rfloor
\end{equation}
where $x,y,z$ represent the location of cloud points, $x_{front}$ and $y_{front}$ indicate the corresponding position on the 2D front view, $Res_{h}$ and $Res_{v}$ are the horizontal and vertical angular resolution in degrees, $d=\sqrt{x^{2}+y^{2}}$ represents the point's distance from the sensor. Since the laser travel time is proportional to the distance, we therefore re-sampled the distance values to fit the cycle time of laser emission, and used this value to construct the pulse array. In order to project the label annotations into the front view, we first calculated the bounding box in camera coordinates and transfer the corners to the Velodyne coordinates by multiplying the transition matrix $T_{c2v}$. The object location can then be mapped into the front view similarly (Figure \ref{fig:kitti_process}). Based on the front view location, different objects were cropped with the fixed size to establish the recognition dataset. It contains 32,456 training samples and 8,000 testing samples covering the 8 classes of KITTI (Figure \ref{fig:kitti}). The `Misc' class contains infrequent objects like trailers and segways.

\subsubsection{DVS Dataset}

% \begin{figure}[ht]
% \centering
% \includegraphics[width=0.6\linewidth]{fig/new1-3.png}
% \caption{DVS\_barrel dataset}
% \label{fig:dvs}
% \end{figure}
Dynamic visual sensor (DVS) is an event-based camera whose pixels work asynchronously and can thus better capture moving objects, especially at high speeds. Thanks to its high measurement rate and low latency compared with standard cameras, DVS has been considered as a very promising sensor in autonomous driving. However, due to the lack of DVS data for autonomous driving, we utilized the DVS\_barrel dataset (Figure \ref{fig:dvs}) with 3,453/3,000 samples for training/testing. DVS encodes the relative changes in illumination asynchronously and generates spike events as the outputs, it therefore fits the SNN model very well.

\subsection{Experimental Setup}

The SNN-based object recognition system with temporal coding was evaluated on the Sim LiDAR, DVS\_barrel and transformed KITTI datasets. For the Sim LiDAR task, a fully connected SNN architecture was adopted, which contains 256/400/30 neurons in the input layer, hidden layer, and output layer respectively. We set the batch size to 60 and trained the SNN using stochastic gradient descent (SGD) with L2 regularization and a learning rate of 1e-2.

A fully connected SNN of larger size was used for the DVS task. It consisted of 1024 input neurons, 2000 hidden ones and 36 output channels. The training employed Adam as optimizer with a batch size of 10 and a learning rate of 1e-3.

As for the KITTI task, considering the complex scenes and backgrounds introduced, an enhanced SCNN was employed. It is composed of two convolutional layers ($sconv2d\_1$ \& $sconv2d\_2$) and two fully connected layers ($sdense\_1$ \& $sdense\_2$). The input size is $50\times118\times1$, the kernel size for $sconv2d\_1$ and $sconv2d\_2$ is $5\times5$, with a stride size of 2. The numbers of kernels are 48 and 24 respectively. The output from $sconv2d\_2$ ($13\times30\times24$) is flattened and passed to $sdense\_1$ (with 256 spiking neurons) and 8 output channels are gained from $sdense\_2$. The batch size was set to 10 and the initial learning rate at 1e-3 with decay. Adam optimizer was adopted for training.

\subsection{Results and Performance Analysis}

With different noise levels as shown in Figure \ref{fig:noise}, the testing accuracy on the Sim LiDAR dataset is shown in Table \ref{tbl:acc_sim}. 
\begin{table}[ht]
\vspace{-5pt}
\caption{Results on Sim LiDAR under different noise levels}
\vspace{-10pt}
\label{tbl:acc_sim}
\begin{center}
\scalebox{0.9}{
\begin{tabular}{l|c|c|c|c}
\toprule
Noise Range & 0 - 0.10 & 0 - 0.20 & 0 - 0.33 & 0 - 0.50  \\ \midrule
Accuracy & 99.83\% & 96.16\% & 82.66\% & 68.16\% \\ \bottomrule
\end{tabular}}
\end{center}
\vspace{-10pt}
\end{table}

\begin{figure}[tbp!]
    \centering
    \includegraphics[width=1\linewidth]{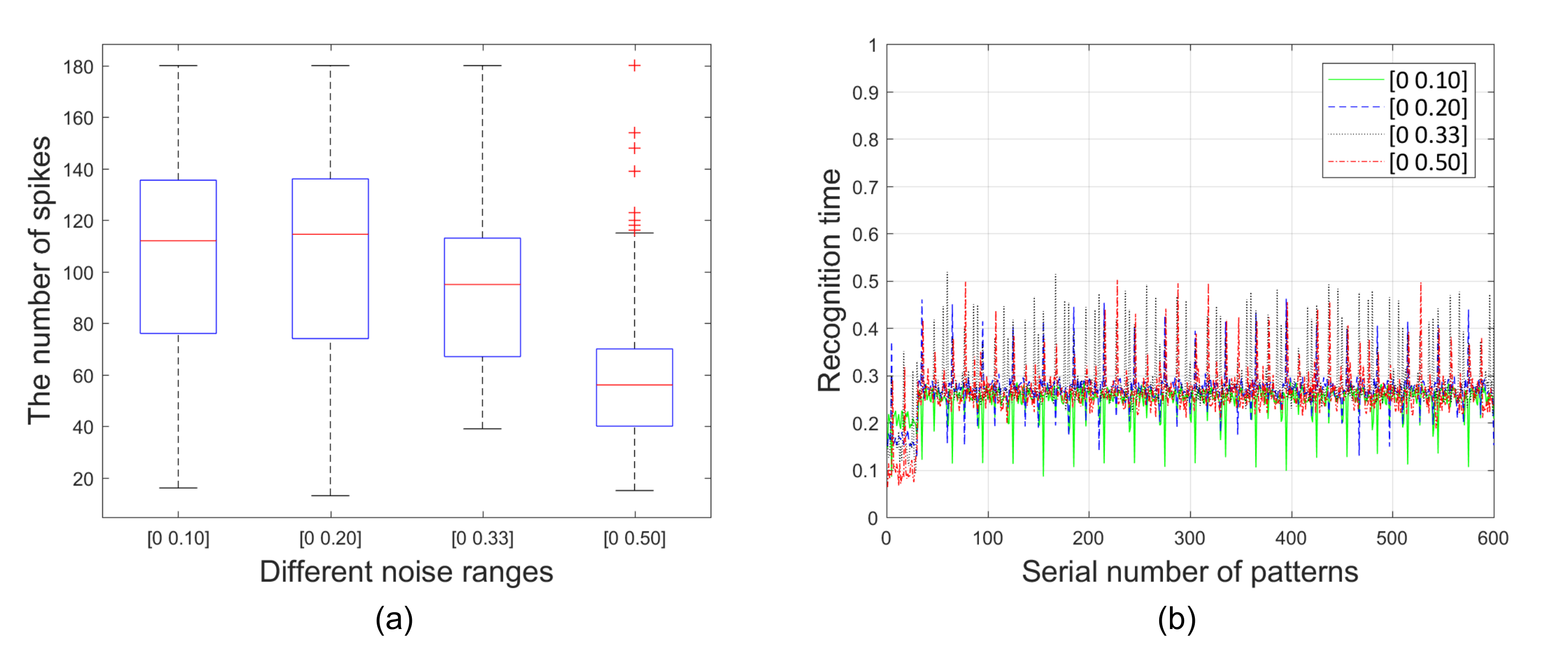}
    \caption{The efficiency of SNN-based object recognition system can be reflected by (a) the number of spikes processed towards decision and (b) inference time}
    \label{fig:effi_sim}
    \vspace{-15pt}
\end{figure}

Figure \ref{fig:effi_sim} shows the recognition performance on the Sim LiDAR dataset. With relatively low noise (0.1 and 0.2), for most cases, it only requires 0.23 ms to 0.3 ms to recognize the object, with the number of spikes as low as 16 and 13 and no exceeding 180, while traditional networks need to process 256 inputs ($16\times16$) to complete the same task. 

For the DVS-barrel dataset, as shown in Table \ref{tbl:comp_barrel}, our model achieved an accuracy of 99.52\%, showing the best performance among all existing models.

\begin{table}[tbp!]
\centering
\caption{Comparison with existing results on DVS-barrel}
\vspace{-10pt}
\label{tbl:comp_barrel}
\scalebox{0.9}{
\begin{tabular}{lrr}
\toprule
Model      & Method   & Acc. \\ \midrule
\cite{perez2013mapping}     & CNN Spike-based     & 91.6\%    \\
\cite{perez2013mapping}     & CNN Frame-based     & 95.2\%    \\
\cite{orchard2015hfirst}    & HFirst Temporal     & 84.9\%    \\
Our model           & SNN                & 99.5\%    \\ \bottomrule
\end{tabular}}
\vspace{-5pt}
\end{table}

% \begin{table}[h]
% \caption{Results on Sim LiDAR under different noise levels}
% \label{tbl:acc_sim}
% \begin{center}
% \scalebox{0.9}{
% \begin{tabular}{c|cccc}
% \hline
% \textbf{Noise Range}    & 0-0.10        & 0-0.20        & 0-0.33        & 0-0.50 \\ \hline
% \textbf{Accuracy}       & 99.83\%\      & 96.16\%\      & 82.66\%\      & 68.16\%\ \\ \hline
% \end{tabular}}
% \end{center}
% \end{table}

% \begin{figure}[ht]
% \centering
% \label{fig:accuracy}
% \includegraphics[width=0.6\linewidth]{fig/accuracy5.png}
% \caption{The average accuracy of test.}
% \end{figure}

Table \ref{tbl:compare_kitti} presents the performance of the proposed SCNN model in comparison with other leading methods on the KITTI dataset, in terms of accuracy, latency and power consumption. Because to our best knowledge, we are the first to use SNN to directly perform object recognition on temporal pulses of LiDAR signals (specifically based on KITTI), and considering that SNN is fundamentally different from conventional NNs, there is hardly an existing method or result for a direct comparison. Other state-of-the-art methods (using CNNs) either address a different task, or work on a different type of data (image frames). In view of this, we aim to provide some performance comparisons by transfer learning and fine-tuning pre-trained VGG-16 and ResNet-50 model on our transformed KITTI dataset (but in an image format). To better evaluate the strength of SNN-based system, a CNN with the same architecture and settings as our SCNN was also trained. 

Besides accuracy, the SNN object recognition system was also evaluated for the recognition time and data points required for recognition (ratio to total), as indicated by $T_{rec}$ and $R_{data}$. Recognition time indicated the time needed for recognition in real-world implementation. It is notable that our proposed model directly consumes pulse data and processes with temporal coding, so both inputs and outputs reflect time information. Therefore, without complex pre-processing of the raw signals, the processing time for recognizing an object can be calculated. For this evaluation, based on the provided Velodyne LiDAR parameters (laser emitting and collecting frequency, around 1-2.2 million points/second), we are able to estimate the processing time for each sample. The number of data points consumed towards a decision represents the count of input values processed by models before the object is recognized. For SNN, it is calculated as the ratio between the number of input spikes consumed before the first output spike and the total number of spikes. Highlighting that, in order to get the recognition result, not all pulses need to be processed. Compared with the conventional NNs that consume all input values for inference, the SNN model can significantly reduce the computational cost and latency.

\begin{table}[tbp!]
\caption{Performance comparison with leading models on KITTI}
\label{tbl:compare_kitti}
\vspace{-10pt}
\begin{center}
\scalebox{0.9}{
\begin{tabular}{l r r r r} 
\toprule
Model   & \begin{tabular}[c]{@{}c@{}}Acc.\\ 
(\%) \end{tabular}      & \begin{tabular}[c]{@{}c@{}}$T_{rec}$\\ 
(ms) \end{tabular}    & \begin{tabular}[c]{@{}c@{}}$R_{data}$\\ 
(\%) \end{tabular}   & \begin{tabular}[c]{@{}c@{}}Power Consumption\\ 
($\alpha$=0.37 pJ / $\alpha$=45 pJ) \end{tabular}\\ 
\midrule
SCNN    & 96.62   & 2.02     & 76    & 29.83 nJ/3.63 $\mu$J\\ 
CNN     & 88.22   & 2.58     & 100    & 0.67 J\\ 
VGG-16  & 92.72      & 11.34      & 100    & 2.95 J\\
ResNet-50 & 92.84      & 71.30      & 100    & 18.54 J\\
\bottomrule
\end{tabular}}
\end{center}
\vspace{-10pt}
\end{table}

% \begin{table}[ht]
% \caption{Comparison between SCNN and CNN on KITTI}
% \label{tbl:compare}
% \begin{center}
% \scalebox{0.9}{
% \begin{tabular}{l|ll}
% \hline
% \textbf{Model}                                      & SCNN             & CNN     \\ \hline
% \textbf{Accuracy}                                   & 96.62\%          & 88.22\% \\
% \textbf{Recog. Time (ms)}                           & 2.027            & 2.583   \\
% \textbf{Number of Spikes}                           & 4512             & 5900    \\
% \textbf{\begin{tabular}[c]{@{}l@{}}Power Consumption\\ 
% ($\alpha$=0.37 pJ / $\alpha$=45 pJ)
% \end{tabular}}                                      & \begin{tabular}[c]{@{}l@{}}29.83 nJ\\/3.63 $\mu$J\end{tabular} & 0.67 J  \\ \hline
% \end{tabular}}
% \end{center}
% \end{table}

It can be seen that our SCNN model gained 96.62\%\ accuracy over all classes, while achieving promising time and computational efficiency. More importantly, it is shown that the SCNN model can work in a real-time and ultra fast manner, requiring only about 2ms on average to recognize an object in the KITTI dataset. Meanwhile, it processes an average of 4512 spikes (5900 as a total) to make a decision, which demonstrates the relatively low computational cost needed.

The CNN achieved 88.22\% accuracy (with same training and testing sets), much lower than the 96.62\% of SCNN, which conforms to the understanding that spiking neurons are fundamentally more powerful computational units than traditional artificial neurons due to the temporal information involved \cite{maass1997networks}. VGG-16 and ResNet are more powerful than CNN but still compromise to noisy sensory data. As for latency, SCNN takes on average 2.027 ms for one sample while CNN, VGG-16 and ResNet take significantly longer. It is notable that the inference times of these three models don't include the time of frame output, which is needless for SNN, so the overall processing time of SNN is much faster.
% are upon the arrival of all data points in a frame while SCNN starts calculation as soon as the first data point comes in. 
In addition, time consumption of SCNN was estimated only based on software simulation. If being deployed on customized hardware, the inference time could be further shortened. 

% The CNN will need all 5900 data points to make a decision, while SCNN only need 4512.3 pulses on average, this makes SCNN faster and more energy efficient. 

\begin{figure}[t]
    \centering
    \includegraphics[width=0.9\linewidth]{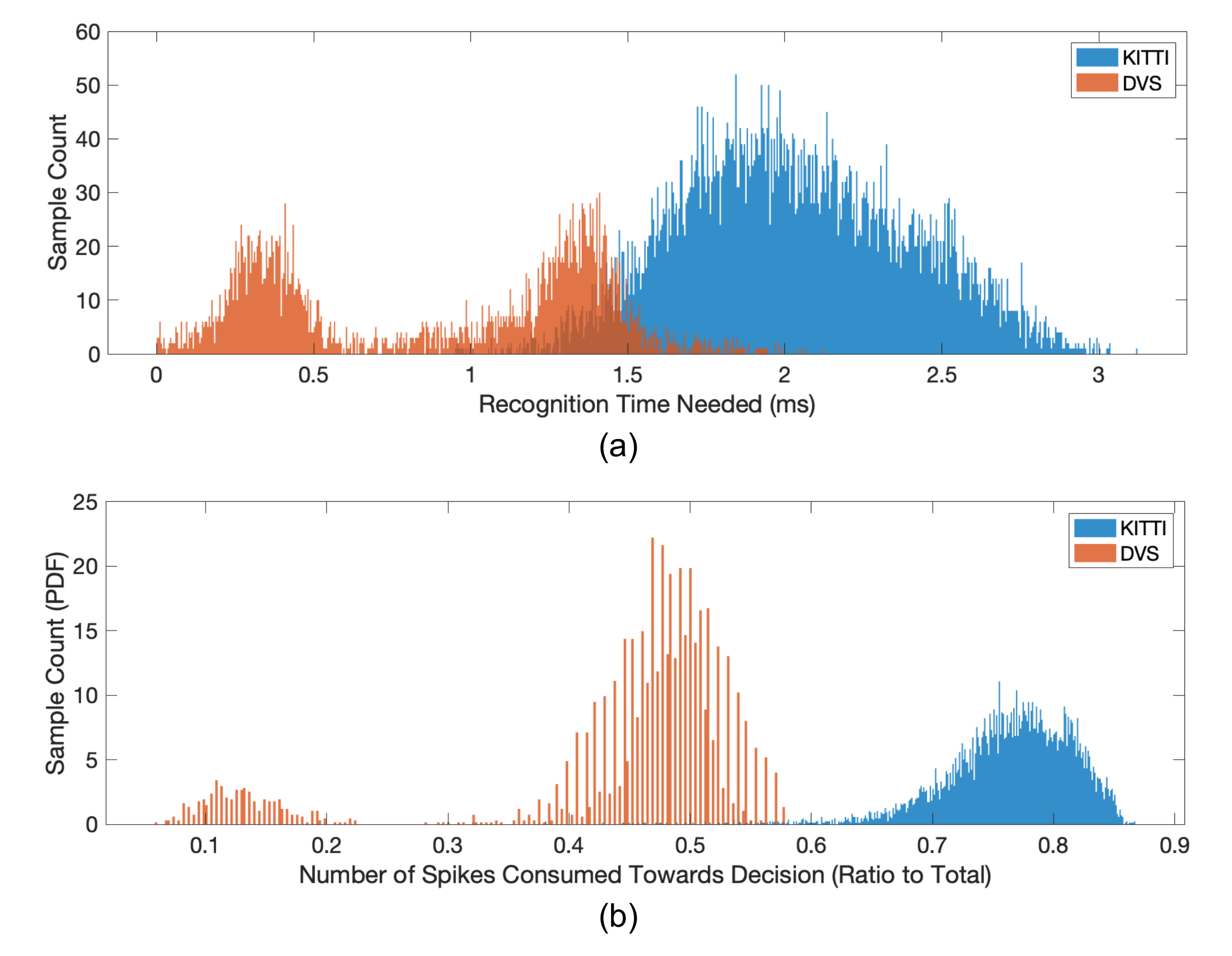}
    \vspace{-10pt}
    \caption{Distribution of (a) recognition time and (b) number of pulses consumed}
    \label{fig:effi_kitti_dvs}
    \vspace{-15pt}
\end{figure}

The efficiency of SNN model can also be manifested from Figure \ref{fig:effi_kitti_dvs}, where the number of spikes needed for KITTI and DVS concentrates in the range 4000-5000 and 100-150 (4512 and 113 spikes on average) respectively, significantly less than traditional networks' 5900 ($50\times118$) and 1024 ($32\times32$). Moreover, the recognition time for KITTI and DVS is mostly less than 3ms and 1.5ms.

Furthermore, we estimated and compared the power consumption between the proposed method and existing approaches. To do so, we adopted the energy analysis methodology in \cite{cao2015spiking} to map SNN to a neuromorphic hardware and made a simplified assumption that a spike activity consumed $\alpha$ Joules of energy. Then for the proposed SNN architecture, we overstated that all spiking neurons were activated, which led to 74,728 spikes, adding the entire input spike sequence 5,900, the maximum energy consumption for recognizing one sample would be 80,628 $\alpha$ Joules. If we apply the power characteristics of two published spike-based neuromorphic circuit \cite{cruz2012energy,merolla2011digital}, where the energy consumption per spike ($\alpha$) are 0.37 pJ and 45 pJ respectively, then the energy consumed by our SCNN for one sample equals 29.83 nJ or 3.63 $\mu$J. Meanwhile, the conventional CNN, VGG-16 and ResNet-50 were trained and tested on NVIDIA GeForce RTX 2080 Ti, with 260 Watt power. According to the average recognition time, the power consumption per sample can be easily calculated, several order of magnitudes higher than both SNN hardware.

% \begin{table}[ht]
% \caption{Performance on KITTI and DVS dataset}
% \label{tbl:acc_kitti_dvs}
% \begin{center}
% \scalebox{0.9}{
% \begin{tabular}{c|c c c} 
% \hline
% \textbf{Task} & \textbf{Accuracy} & \textbf{Recog. Time (ms)} & \textbf{Num. Spikes}\\ \hline
% KITTI            & 96.62\%           & 2.0273           & 4512.3\\ 
% DVS              & 99.52\%           & 0.9044           & 113.1 \\ \hline
% \end{tabular}}
% \end{center}
% \end{table}

% The column \textit{``Recog. Time''} shows the estimated average 

Based on the above results, the proposed SNN approach can achieve admirable computational and time efficiency.

\section{Conclusion}
\label{sec:conclusion}

In this paper we proposed an SNN-based object recognition system utilizing temporal coding. To the best of our knowledge, this is the first SNN model that directly processes LiDAR temporal pulse signals for object recognition in autonomous driving settings. A standard feedforward architecture and an extended SCNN were proposed and evaluated based on three representative datasets, Sim LiDAR, KITTI and DVS. The performance results have proved that, the proposed system can achieve remarkable accuracy on real-word data and significantly reduce the computational cost while working in a real-time manner when being deployed on hardware. It demonstrates the potential of SNN in autonomous driving and other resource-/time-constrained applications. Future investigation will be made to build SNN models for effective and efficient object detection.

% \begin{table}
% \centering
% \begin{tabular}{lrr}  
% \toprule
% Scenario  & $\delta$ (s) & Runtime (ms) \\
% \midrule
% Paris       & 0.1  & 13.65      \\
%             & 0.2  & 0.01       \\
% New York    & 0.1  & 92.50      \\
% Singapore   & 0.1  & 33.33      \\
%             & 0.2  & 23.01      \\
% \bottomrule
% \end{tabular}
% \caption{Booktabs table}
% \label{tab:booktabs}
% \end{table}

\newpage

\newpage
\bibliographystyle{named}
\bibliography{ijcai20}

\end{document}